\documentclass{article}

\usepackage{graphicx}
\usepackage{microtype}
\usepackage{caption}
\usepackage[latin1]{inputenc}
\usepackage{amssymb,amsmath,array}
\usepackage{amsthm}
\usepackage{epsfig}
\usepackage{url}
\usepackage{subcaption}
\usepackage[algo2e,linesnumbered,boxed]{algorithm2e} 
\usepackage[hidelinks]{hyperref}
\newcommand{\typicalfig}[3]
 {
	\begin{figure}[bt]
		\centering		
			\includegraphics[width=.81\columnwidth]{#1}
					\caption{#2}
		\label{fig:#3}
	\end{figure}
	}
\newcommand{\stickfig}[3]
{
\begin{figure}[bt]
	\centering
		\includegraphics[width=.6\columnwidth]{#1}
	\caption{Stick figure of different body parts related to the \textit{#2 dataset}. Red markers are the selected important inputs according to the regularised relevance profile of the features.}
	\label{fig:#2_stick}
\end{figure}
}
\newcommand{\featselfig}[3]
{
\begin{figure}[bt]
	\centering
		\includegraphics[width=.81\columnwidth]{#1}
	\caption{Classification performance of \textit{#2 dataset} based on the selected features according to the regularised profile  }
	\label{fig:#2_feat_sel}
\end{figure}
}
\newcommand{\relfig}[2]
 {
	\begin{figure}[bt]
		\centering		
			\includegraphics[width=.81\columnwidth]{#1}
			\caption{Average (blue bars) and deviation (red lines) of  the relevance values for features of the \textit{#2 dataset} calculated according to the normalised diagonal values of $(\mathbf{L}^t\mathbf{L})$. \textit{Top:} Regularised relevance profile. \textit{Bottom:} Non-regularised relevance profile.}
		\label{fig:#2_Rel}
	\end{figure}
	}

\hypersetup{
	pdfstartview	= {FitH},
	pdftitle			= {Efficient Metric Learning for the Analysis of Motion Data},
	pdfsubject		= {ESANN 2018},
	pdfauthor 		= {Babak Hosseini and Barbara Hammer},
	pdfcreator 		= {XeLaTeX},
	pdfproducer		= {LaTeX with hyperref}
}
\urldef{\mailsa}\path|{bhosseini, bhammer}@techfak.uni-bielefeld.de|

\theoremstyle{proposition}

\theoremstyle{theorem}

\theoremstyle{definition}

\theoremstyle{lemma}

\theoremstyle{remark}

\SetKwInput{KwInput}{Input}
\SetKwInput{KwOutput}{Output}
\SetKwInput{KwTask}{Task}
\SetKwInput{KwInit}{Initialization}
\SetKwInput{KwProc}{Loop}
\SetKwInput{KwStepk}{Step K}
\date{\footnotesize{Preprint of the publication~\cite{hosseini2015efficient}, as provided by the authors.\\
		The final publication is available via \url{https://ieeexplore.ieee.org/document/7344819} }}
\begin{document}
	%style file for ESANN manuscripts
	\title{Efficient Metric Learning\\ for the Analysis of Motion Data}
	
	%***********************************************************************
	% AUTHORS INFORMATION AREA
	%***********************************************************************
	\author{Babak Hosseini$^1$ and Barbara Hammer$^1$
		%
		% Optional short acknowledgment: remove next line if non-needed
		\thanks{This research was supported by the Cluster of Excellence Cognitive 
			Interaction Technology 'CITEC' (EXC 277) at Bielefeld University, which
			is funded by the German Research Foundation (DFG).}
		%
		% DO NOT MODIFY THE FOLLOWING '\vspace' ARGUMENT
		\vspace{.3cm}\\
		%
		% Addresses and institutions (remove "1- " in case of a single institution)
		CITEC centre of excellence, Bielefeld University\\
		Bielefeld, Germany\\
		%{bhosseini,bhammer\}@techfak.uni-bielefeld.de
	}
	%***********************************************************************
	% END OF AUTHORS INFORMATION AREA
	%***********************************************************************
	
	\maketitle
	
	\begin{abstract}
		We investigate metric learning in the context of dynamic time warping (DTW), 
		the by far most popular dissimilarity measure used for the comparison and analysis of motion capture data. 
		While metric learning enables a problem-adapted representation of data, the majority of methods has been proposed 
		for vectorial data only. In this contribution, we  extend the popular principle offered by the large margin nearest neighbors learner (LMNN) to  DTW by treating the resulting component-wise dissimilarity values as features.
		We demonstrate that this principle greatly enhances the classification accuracy in several benchmarks.
		Further, we show that recent auxiliary concepts such as metric regularization can be transferred from the vectorial case to component-wise DTW in a similar way. 
		We illustrate that metric regularization constitutes a crucial prerequisite for the interpretation of the resulting relevance profiles.
	\end{abstract}
	
	\section{Introduction}
Motion capture (Mocap) systems rely on a variety of different principles such
as magnetic or mechanical sensors, optical markers, poseable mannequins, dedicated technology for hand or facial expression tracking, and low-cost marker-less technology
\cite{markerless,DBLP:journals/pami/LiuGSDST13}.
Powerful analysis software enables the reconstruction of the underlying
skeleton when dealing with human motion 
\cite{kinematics,animation,DBLP:series/acvpr/978-1-4471-4640-7}.
These developments cause an  increasingly important role
of Mocap in diverse areas such as entertainment, 
sports, or medical applications. 
When Mocap information is used in complex systems such as virtual trainers or
interactive Mocap databases for medical analysis, intelligent data analysis, and
machine learning technology become necessary.
Proposed methods range from the independent component analysis (ICA)
up to deep learning \cite{modeep,BurgerToiviainen2013}.

Within such systems, distance-based methods are often used 
for the initial analysis or motion retrieval 
\cite{6460780,retrieval,DBLP:conf/ecir/DemuthRME06}.
Due to its capability of adjusting to
different durations, dynamic time warping (DTW) constitutes the by far most popular 
dissimilarity measure in this context
\cite{Zhou_2012_7021,4665151,DBLP:conf/icdm/PetitjeanFWNCK14}.

Like any other dissimilarity measure or metric,  the results of DTW
crucially rely on the choice of its intrinsic parameters; 
Crucial metric parameters for DTW and similar
measures such as alignment are the parameters which determine how two single sequence entries are compared.
This is often referred to as the scoring matrix provided a discrete
alphabet is chosen, such as DNA or protein sequences.
When it comes to sequence alignment in bioinformatics,
lots of effort has been made for its correct
choice  based on biological insight \cite{citeulike:2492411}.
Provided such insight is not always available, so-called inverse alignment can help
to infer metric parameters from given, optimum alignments
\cite{DBLP:journals/bmcbi/Edgar09}\cite{Boyer2008672}.
In general, no such information is present, rather only weak learning signals such as
 motion labeling or grouping are available.
In such cases, the adaptation of metric parameters can be solved by metric learning within a machine learning framework.

Metric learning constitutes a matured field of research 
for the standard vectorial setting (data represented by feature vectors), e.g., see~ 
\cite{Bellet2013,Kulis2013,gmlvq,lmnn}.
In these approaches, usually, quadratic forms are inferred from the given
auxiliary information. This vectorial metric adaptation does not only provide
increased model accuracy, but it also dramatically facilitates model
interpretability, and it can lead to additional functionalities such as 
a direct data visualization 
\cite{Biehl2013,Backhaus2014,NN12Bunetal}. 
Recently, some researches have focussed on the validity of the 
interpretation of metric parameters as relevance weights, and it has been shown
that there exist problems in particular for high-dimensional or
highly correlated data \cite{l1reg,l2reg}. It is possible to avoid these
problems by an efficient form of metric regularization
as detailed in the approaches\cite{l1reg,l2reg}.

These developments mainly focus on the context of vectorial data;
therefore, they are not applicable to distance-based measures, e.g, in the case of DTW. Recently, a few
approaches have been proposed which address metric-parameter
learning for complex non-vectorial data, in particular
sequences and sequence alignment
\cite{Bernard20082611,Bellet2013,2710031}.
While these approaches lead to an increased
model accuracy and interpretability, 
they have the drawback that their training complexity is very costly:
typically, these techniques adapt metric parameters within sequence alignment,
such that pairwise distances of all data samples have to be recomputed after
every metric adaptation step.

In this contribution, we take a different point of view for the sake of a
significantly reduced computational load: we rely on a representation
of sequential data in terms of pairwise dissimilarity vectors as provided
by component-wise DTW. This strategy is similar to the popular treatment of dissimilarity data
as \lq features\rq, which is detailed in the monograph 
\cite{diss}. This mathematical formulation enables us to transfer 
the powerful 
large margin nearest neighbors (LMNN) metric learner \cite{lmnn} to a metric adaptation for DTW,
which is able to adjust the relevance of single joints and their correlations
in the Mocap data according to a given specific classification task. Further, we demonstrate
that a recent metric regularization framework can be transferred to Mocap DTW
results
based on the same principle, this way guaranteeing a valid interpretation of the resulting relevance profiles.
We demonstrate the efficiency and effectiveness of the proposed methodology for different
benchmark datasets.

This contribution is structured as follows:
first, we introduce LMNN and its transfer to component-wise DTW, dubbed DTW-LMNN.
Afterward, we discuss how the resulting metric can be regularized to avoid 
its dependency on random effects caused by correlations in the observed data.
Finally, we demonstrate the suitability of the proposed methodology in a couple
of benchmark Mocap data, where we demonstrate the increased classification
accuracy of DTW-LMNN in comparison to alternatives. Finally, we demonstrate the
increased interpretability and robustness of the results when metric regularization
takes place.

%%%%%%%%%%%%%%%%%%%%%%%%%%%%%%%%%%%%%%%%%%% LMNN
%\clearpage
\section{Dissimilarity based metric learning}
\label{sec:lmnn}
In this section, we review the LMNN algorithm and the DTW distance computation. Then we introduce the proposed algorithm to use the DTW metric along with metric learning optimization,
and we discuss how to regularise the resulting dissimilarities to diminish random effects caused
by data correlations. Note that DTW is not a metric since the triangle inequality does not hold, 
rather it is a pairwise symmetric distance function, which can serve as a data dissimilarity measure. Nevertheless, 
we carelessly refer to DTW as a metric in some places, although strong metric properties do not apply.

\subsection{Large margin nearest neighbors metric learning}
LMNN is a metric learning algorithm which learns a quadratic form from given 
labeled data $(\vec x^i,y_i)\in\mathbb{R}^n\times\{1,\ldots,c\}$, $i=1,\ldots, m$,
$c=$ number of classes,
to improve the classification accuracy of the well-known $k$-nearest neighbors (KNN) method. 
As a distance-based approach, the accuracy of KNN fundamentally relies on its underlying 
distance measure which defines the $k$ nearest neighbors of a given data point.
LMNN tries to robustly adjust this neighborhood structure by learning a parameterized form
%$^\top$ or $^\intercal$
\begin{equation}
{\cal D}(\vec x^i,\vec x^j)=(\mathbf{L}(\vec x^i-\vec x^j))^2 = (\vec x^i-\vec x^j)^\top\mathbf{L}^\top\mathbf{L}(\vec x^i-\vec x^j)
\label{metric}
\end{equation}
with adjustable linear transformation matrix $\mathbf{L}$ which
induces a quadratic form characterized by $\mathbf{M}:=\mathbf{L}^\top\mathbf{L}$.
The objective function of LMNN is based on a fixed $k$-neighborhood structure
characterized by $\eta_{ij}\in\{0,1\}$ where $\eta_{ij}=1$ iff
point $\vec x^j$ is within the $k$ closest neighbors of $\vec x^i$.
Based on the intuition of having dense neighborhoods, while maximizing distances of
a data point to its neighbors with different labeling, the costs of LMNN become
\begin{equation}
\begin{array}{l}
\epsilon({\mathbf L})  := 
\sum_{ij}\eta_{ij}{\cal D}(\vec x^i,\vec x^j)
+ c \sum_{ijl}\eta_{ij}(1-\delta_{y_i}^{y_l})\cdot\\
\left[1+{\cal D}(\vec x^i,\vec x^j)-{\cal D}(\vec x^i,\vec x^l)\right]_+
\label{lmnncosts}
\end{array}\end{equation}
where $[\cdot]_+$ refers to the Hinge loss
and $c>0$ is an adjustable meta parameter.
This objective can be interpreted as the goal to adjust the metric
such that all points with different class labels
are located outside of the data neighborhood with a margin.
It has been shown in \cite{lmnn} that this optimization problem is equivalent to the following
semi-definite optimization:
\begin{equation}
\begin{array}{ll}
\mathrm{min}  &\sum_{ij} {\cal D}(\vec x^i,\vec x^j) + c\sum_{ijl}\eta_{ij}(1-\delta_{y_i}^{y_l})\xi_{ijl}\\
\mathrm{where} & {\cal D}(\vec x^i,\vec x^l) - {\cal D}(\vec x^i,\vec x^j) \ge 1-\xi_{ijl}\\
&\xi_{ijl}\ge 0\\
& \mathbf{M}\ge 0
\label{lmnn}
\end{array}
\end{equation}
which can be optimized efficiently w.r.t.\ the matrix $\mathbf{M}$.
Note that it is possible to choose a low-rank matrix $\mathbf{M}$ which corresponds to
a low-dimensional projection $\mathbf{L}$ of the data vectors.

\subsection{Dynamic time warping}
Rather than vectors, we deal with sequential data
$X^i=(\vec x^i(1) \ldots\,\vec x^i(T)) \in (\mathbb{R}^n)^*$ 
where $T$ denotes the length of the time series.
DTW aligns two time series of possibly different lengths according to warping paths such
that the aligned points match as much as possible, respecting the temporal ordering of
the sequence entries.
Dynamic programming enables efficient computation of
an optimum match in quadratic time with respect to sequence lengths.
For the exact formulas as well as ways to speed up
the computation, we refer to \cite{dtw}.
The following facts are of interest to us:

\noindent
{\bf (I)} Given two time series of possibly different length $X^i$ and $X^j$,
DTW provides a dissimilarity 
${\cal D}_{\mathrm{DTW}}(X^i,X^j)$. 

\noindent
{\bf (II)}
There exist two different ways to treat the vectorial nature of the sequence entries:

\noindent
{\bf (II.1)}
We can directly compute DTW on vectorial
sequences. Then, the outcome of DTW is determined by choosing the parameters of the metric which is used to
compare vectorial sequence entries along the warping path.
meaning that crucial metric parameters are those involved in
computing
${\cal D}(\vec x^i(t_1),\vec x^j(t_2))$
where the warping path determines the time points
$(t_1,t_2)$,
and ${\cal D}:\mathbb{R}^n\times\mathbb{R}^n\to\mathbb{R}$ is a vectorial metric used to compare the vectorial sequences.

\noindent
{\bf (II.2)}
We can compute DTW separately for every dimension of a given time series 
$X^i_k = (x_k^i(1) \ldots\, x_k^i(T)) \in\mathbb{R}^*$,
where $k\in\{1,\dots,n\}$ refers to the component $k$ of the vectorial sequence entries.
For two time series, we thus get a vector of distances
\begin{equation}
\vec{D}^{ij}:=({\cal D}_{\mathrm{DTW}}(X^i_1,X^j_1),\ldots,{\cal D}_{\mathrm{DTW}}(X^i_n,X^j_n))
\in\mathbb{R}^n
\label{distances}
\end{equation}
of dimensionality $n$.
A real-valued dissimilarity can be computed thereof
by a standard quadratic form:
\begin{equation}
\begin{array}{l}
{\cal D}_{\mathrm{LMNN-DTW}}(X^i,X^j) := ({\mathbf L}\cdot\vec D^{ij})^2=\\
\left(\mathbf{L}\cdot({\cal D}_{\mathrm{DTW}}(X^i_1,X^j_1),\ldots,{\cal D}_{\mathrm{DTW}}(X^i_n,X^j_n))\right)^2
\end{array}
\label{dtwmetric}
\end{equation}
which is parameterized by a linear mapping $\mathbf{L}:\mathbb{R}^n\to\mathbb{R}^n$ 
(or a low-dimensional counterpart $\mathbf{L}:\mathbb{R}^n\to\mathbb{R}^{n'}$ where $n'<n$).
In both cases, metric parameters are 
in the form of a linear
transformation $\mathbf{L}$ or corresponding
quadratic matrix $\mathbf{M}=\mathbf{L}^\top\mathbf{L}$, which
have to be adapted according to the given problem for an optimal result.
Recently, a few approaches have been proposed which
deal with the question how to learn an optimum transformation
$\mathbf{L}$ provided DTW is used for vectors
(e.g., regarding {\bf (II.1)}), e.g., see~\cite{Bernard20082611,Bellet2013,2710031}.
These techniques, however, face the problem that metric adaptation
can change the form of an optimum warping path, e.g.,
a computationally costly recalculation of the warping path is necessary to obtain stable
results.

Therefore, we propose an approach in the following which is based
on the strategy {\bf (II.2)} to deal with vectorial data:
we adapt the metric according to component-wise DTW vectors (\ref{dtwmetric}).
This formulation has the benefit that not only LMNN can efficiently be transferred
to a novel dissimilarity-based setting, but also recent concepts for metric 
regularization apply to such problems, as we see in the following.

\subsection{DTW-LMNN}
For a sequence metric such as
${\cal D}_{\mathrm{LMNN-DTW}}$,   the LMNN costs (\ref{lmnncosts}) become:
\begin{equation}
\begin{array}{l}
\epsilon({\mathbf L})  = 
\sum_{ij}\eta_{ij}{\cal D}_{\mathrm{LMNN-DTW}}(X^i,X^j)
+ c \sum_{ijl}\eta_{ij}(1-\delta_{y_i}^{y_l})\\
\left[1+{\cal D}_{\mathrm{LMNN-DTW}}(X^i,X^j)
\mbox{}-{\cal D}_{\mathrm{LMNN-DTW}}(X^i,X^l)\right]_+.
\label{dtwlmnncosts}
\end{array}\end{equation}
Using the distance computation (Eq.~\ref{dtwmetric}), we obtain 
an optimization problem which is similar to (Eq.~\ref{lmnn}):
\begin{equation}
\begin{array}{ll}
\mathrm{min}  &\sum_{ij} (\vec D^{ij})^\top {\mathbf M} \vec D^{ij}  + c\sum_{ijl}\eta_{ij}(1-\delta_{y_i}^{y_l})\xi_{ijl}\\
\mathrm{where} & (\vec D^{il})^\top {\mathbf M} \vec D^{il}  - (\vec D^{ij})^\top {\mathbf M} \vec D^{ij}  \ge 1-\xi_{ijl}\\
&\xi_{ijl}\ge 0\\
& \mathbf{M}\ge 0.
\label{dtwlmnn}
\end{array}
\end{equation}
This problem can be solved by means of semi-definite programming.
Again, a restriction to low-rank matrices $\mathbf{M}$ and $\mathbf{L}$ is possible,
provided the relevant information is located in a low-rank subspace of 
the full data space only.
Note that the computational complexity of this method is the same as for
vectorial LMNN; further, the convexity of the problem is preserved.

\subsection{Metric regularization}
The adaptation of a quadratic form as present in LMNN does not only
enhance the classification accuracy, but it can also give rise to increased
interpretability of the results. A quadratic form corresponds to the linear data
transformation ${\vec x^i \mapsto \mathbf{L}}\vec x^i$.
Hence the diagonal terms of the matrix $\mathbf M$
\begin{equation}
M_{kk} =  \sum_i L_{ik}^2
\label{relevance}
\end{equation}
summarise the influence of feature $k$ on the mapping.
Due to this observation, metric learners are often accompanied by
the \textit{relevance profile} which is provided by the diagonal entries of ${\mathbf M}$;
this gives insight into relevant
features for the given task, such as potential biomarkers for medical
diagnostics \cite{arltbiehl2011jcem}.
For DTW-LMNN, this interpretation directly transfers to a
relevance profile for the sequential data related to each feature
component, such as single joints in the case of Mocap data:
for the metric (Eq.~\ref{dtwmetric}), the entry $M_{kk}$ summarises the influence
of distances which are based on the feature sensor $k$.

It has recently been pointed out that this interpretation has problems provided high
dimensional or highly correlated data are analyzed: 
in such cases, the relevance profile
and the underlying linear transformation $\mathbf{L}$ are not unique,
rather data correlations can give rise to random, spurious relevance peaks.
 We expect this effect for Mocap
data due to a high correlation of neighboring joints. 
For vectorial data, this effect is caused by the following observation, as pointed
out in \cite{l2reg}:
assume $\mathbf{X}=[\vec x^1,\ldots,\vec x^m]$ refers to the data matrix.
Then two linear transformations $\mathbf{L}_1$ and $\mathbf{L}_2$ are equivalent
with respect to $\mathbf{X}$ iff
$\mathbf{L}_1\mathbf{X} = \mathbf{L}_2\mathbf{X}$.
This relationship is equivalent to the fact that the difference
$(\mathbf{L}_1-\mathbf{L}_2)\mathbf{X}$ vanishes.
Hence, by considering the squared form
\begin{equation}
(\mathbf{L}_1-\mathbf{L}_2)\mathbf{XX}^\top(\mathbf{L}_1-\mathbf{L}_2)^\top =0,
\label{relsquare}
\end{equation}
we can relate this property to the fact that the
differences of the rows are given by vectors which lie in the null space
of the data correlation matrix $\mathbf{C}:=\mathbf{X}\mathbf{X}^\top$.
This fact gives us a unique characterization of the equivalence class
of matrix $\mathbf{L}$ with respect to the data transformations for $\mathbf{X}$:
equivalent matrices, e.g.,
matrices which map data  $\mathbf{X}$
in the same way as matrix $\mathbf{L}$,
differ from $\mathbf{L}$ by multiples of eigenvectors related to $0$ eigenvalues of $\mathbf{C}$. 
Provided the metric learning method does not take this fact into account, its
outcome matrix is random as concerns contributions of this null space.

For LMNN, costs are invariant to null space contributions, e.g.,
the matrix $\mathbf{L}$ is random in this respect. Albeit this property does not affect the training data $\mathbf{X}$,
it influences the result in two aspects:
for test data, the null space is usually different, e.g., the generalization ability
of the model is affected by random effects of the  training data correlation 
and the initialization point $\mathbf{L}$ for the optimization problem.
Second, more severely, random contributions of the null space
of $\mathbf{C}$ change the relevance profile $M_{kk}$
and can give rise to spurious effects such as high values which are not supported by
any \lq real\rq\ relevance of the feature $k$.

Therefore, it is advisable to regularise the matrix $\mathbf{L}$ by relying on the
representative of the equivalence class of $\mathbf{L}$
with the smallest Frobenius norm. 
Equivalently we can consider a projection of $\mathbf{L}$ to the space of eigenvectors of $\mathbf{C}$ with non-vanishing
eigenvalues, or more precisely, the unique transformation
\begin{equation}
\begin{array}{l}
\tilde{\mathbf{L}} := \mathbf{L}\Phi\\
\mbox{where }\Phi := \sum_{j=1}^J \vec u^j(\vec u^j)^\top
\mbox{ with the eigenvectors }\\ \vec u^1, \ldots, \vec u^j \mbox{ of } \mathbf{C} \mbox{ with nonvanishing eigenvalues}.
\end{array}
\label{trafo}
\end{equation}
For vectorial data, the same effect can be obtained by deleting the null space from the data vectors in the first place employing the principal component analysis (PCA), as a very popular preprocessing approach.
However, the reformulation as matrix regularization has the benefit that this 
principle can directly be transferred to more general data such as the alignment vectors
$\vec D^{ij}$, as we see in the following.

For alignment vectors (Eq.~\ref{distances}) and the distance (Eq.~\ref{dtwmetric}), we find
\begin{equation}
\begin{array}{l}
\mathbf{L}_1\vec D^{ij} = \mathbf{L}_2\vec D^{ij} \mbox{ for all } i, j\\
\iff (\mathbf{L}_1-\mathbf{L}_2) \vec D^{ij} = 0 \mbox{ for all } i,j.
\end{array}
\label{alignmentinvariance}
\end{equation}
Hence, similar to (Eq.~\ref{relsquare}), transformations are equivalent with respect to the given data
iff their difference lies in the null space of the correlation matrix $\mathbf{D}\mathbf{D}^\top$
for the distance matrix 
$\mathbf{D} := [\vec D^{11}, \ldots, \vec D^{1m},\ldots, \vec D^{m1}, \ldots, \vec D^{mm}]$,
consisting of all $n$-dimensional vectors of pairwise distances.
Note that this observation enables an effective regularization of the matrix
$\mathbf{L}$ (and $\mathbf{M}=\mathbf{L}^\top\mathbf{L}$) in the same way as for
the vectorial case, relying on the regularization (Eq.~\ref{trafo}):
\begin{equation}
\begin{array}{l}
\tilde{\mathbf{L}} := \mathbf{L}\Phi\\
\mbox{where }\Phi = \sum_{j=1}^J \vec u^j(\vec u^j)^\top
\mbox{ with the eigenvectors }\\ \vec u^1, \ldots, \vec u^j \mbox{ of } \mathbf{D}\mathbf{D}^\top \mbox{ with nonvanishing eigenvalues}.
\end{array}
\label{dtwtrafo}
\end{equation}
As for the vectorial case, this yields the equivalent matrix $\tilde{\mathbf{L}}$
of $\mathbf{L}$ with the smallest Frobenius norm,
for which an interpretation of the diagonal entries becomes possible.
Thereby, this principle is applicable for full-rank matrices as well as low-rank counterparts.
We see in the experiments section that matrix regularization has a substantial effect on the
variance of the resulting relevance profile. Further, it can also enable a 
slightly better generalization ability since it suppresses noise in the given data.

	%%%%%%%%%%%%%%%%%%%%%%%%%%%%%%%%%%%%%%%%%%% Data description

\section{Datasets and Experiments}\label{sec:dataset}
We compare DTW-LMNN to alternatives with a focus on two different 
aspects: its classification accuracy when used within a KNN method,
and its capability to lead to dimensionality reduction by a reduced relevance profile. 
In this section, we first specify the different training models which are compared to
the method LMNN-DTW. Afterward, we explain the used benchmark data.
Experiments are conducted in three steps for these data: first, we compare
the classification accuracy of the proposed method.
Then we investigate its low-rank counterparts. Finally, we discuss the effect
of regularization on the obtained relevance profiles.
The code of DTW-LMNN algorithm is available in an online public repository\footnote{https://github.com/bab-git/dist-LMNN}.

\subsection{Methods}

\noindent
We use the following  main pipelines for comparison:
\begin{itemize}
\item{\textbf{DTW-LMNN:}}
We use component-wise DTW (\ref{distances}) together with metric learning by 
LMNN (\ref{dtwlmnncosts}). Based on the found distances (\ref{dtwmetric}),
a KNN classifier with $k=3$ is evaluated.
We use this technique with a full-rank matrix, as well as a low-rank
matrix with rank $3$, to investigate LMNN's ability to infer a low-dimensional 
representation of the data with high accuracy.
\item{\textbf{DTW-KNN:}} 
We use the plain DTW distance together with a KNN classifier,
without any metric adaptation. Again, $k=3$.
This setting can serve as a baseline to evaluate the improvement
obtained by metric learning for sequential data.
\item{\textbf{Euclidean LMNN:}}
As the second comparison, we use LMNN based on the
standard Euclidean metric instead of DTW, together with a subsequent KNN classifier with $k=3$.
More precisely, we use the LMNN formulation (\ref{dtwlmnn}) with the choice
\begin{equation}
\vec{D}^{ij}:=({\cal D}_{\mathrm{Euc}}(\vec X^i_1,\vec X^j_1),\ldots,{\cal D}_{\mathrm{Euc}}(\vec X^i_n,\vec X^j_n))
\in\mathbb{R}^n
\label{eucdistances}
\end{equation}
where ${\cal D}_{\mathrm{Euc}}$ refers to the standard Euclidean distance
of the vectors $\vec X^i_k$.
One problem consists in the fact that  the considered
time series $X^i$ and $X^j$ have different length, and we have to compute
a vectorial representation $\vec X^i_k$ and $\vec X^j_k$  of equal dimensionality for every component
$k$. Hence, we select the entries $\vec X^i_k:=(x^i_k(1),\ldots x^i_k(T))$ for the shorter time series,
and we subsample $T$ entries $(x^j_k(t_1),\ldots x^j_k(t_T))$ of $X^j$ at equal time intervals 
$1=t_1<\ldots<t_T$ to turn the second
time series into vectors of the same dimensionality.
This setting allows us to judge the effect of DTW as compared
to the standard Euclidean distance, equipped with metric learning.
Again, we investigate this setting with a full-rank and a low-rank adapted matrix with rank $3$.

\begin{figure}[tb]%
	\centering
	\includegraphics[width=0.6\columnwidth]{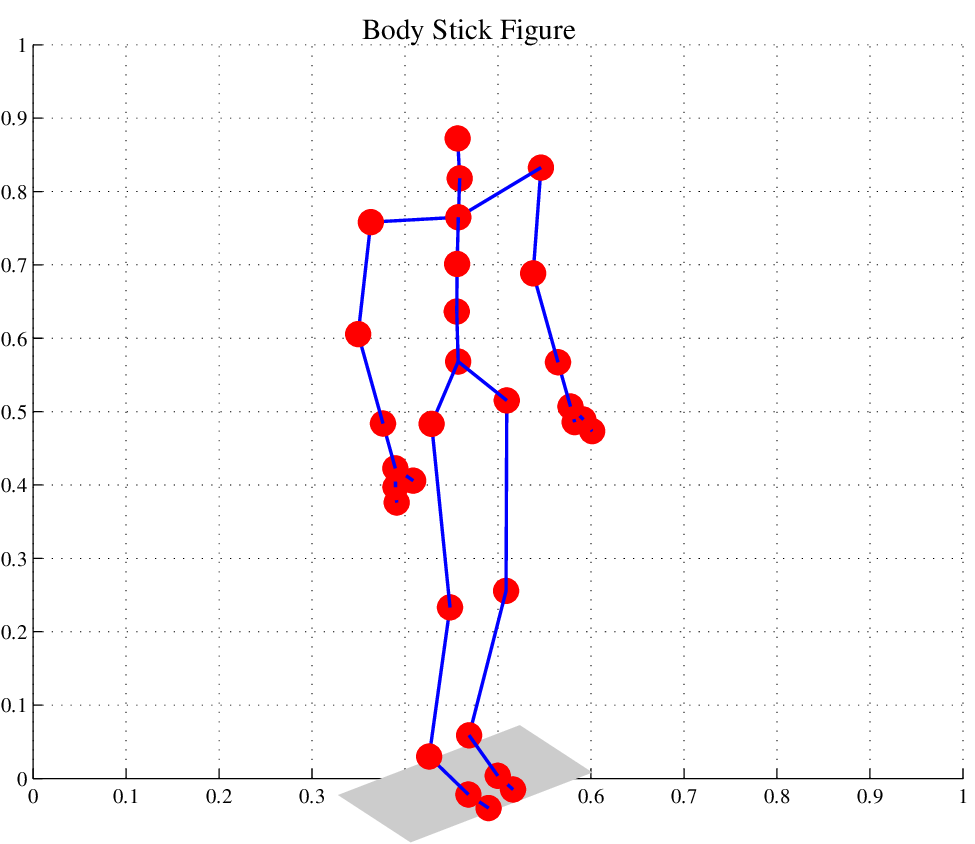}%
	\caption{A stick-figure of the human body showing $41$ markers (red spheres) placed on different parts of 
	the body in order to capture the motion data. Captured data are related to movements of 
	the different joints during the human's basic activities.}%
	\label{fig:stick_figure}%
\end{figure} 

\item{\textbf{PCA-DTW-KNN:}}
For the dimension reduction experiment in section \ref{sec:sec_DR}, we compare the low-rank representation
found by low-rank LMNN with a dimensionality reduction by classical  Principle Component Analysis (PCA) \cite{jolliffe2002principal}. Thereby, PCA is applied directly to the vectorial sequence entries from the training set.
Afterward, DTW is applied on the first 3 principal components of the data,
followed by a KNN classifier. 
\end{itemize}
We use $10$-fold cross-validations with $10$ repetitions for evaluation.
All experiments are carried out using the same cross-validation partitions. 

For evaluation and comparison of the proposed approach, we consider $3$ different Mocap databases based on human motions. These data give rise to the following four different training sets:

\subsection{CMU Mocap dataset:}
We use the Human motion capture dataset from the CMU graphics laboratory \cite{CMU_mocap}. 
The data is captured by Vicon infra-red cameras using $41$ markers placed at different parts of the body, 
close to the joints (Fig.\ref{fig:stick_figure}). Afterward, the images 
are augmented to 3D data and transferred to kinematic information
such as joint rotation/translation based on skeleton information. 
The result consists of $62$ body features leading to $62$-dimensional time series with a high correlation of
their dimensions.  
We select two classification scenarios from the full data set 
to investigate different aspects of the proposed method.

\begin{itemize}
\item{\textit{Walking:}}
This dataset contains data from $7$ different walking styles (normal, fast, slow, turn right, turn left, veer right and veer left) carried out by $4$ different subjects. the dataset consists of $49$ samples ($7$ samples per class) with a $62$-dimensional representation.
\item{\textit{Dance:}}
We use $35$ samples of data related to two different styles of dancing: \emph{Modern} and \emph{Indian}; these are 
available from CMU graphics laboratory as participant subjects $5$ and $94$ \cite{CMU_mocap}. Each class 
contains various dance performances related to subcategories of the class; as a result, the variations 
among the data within the classes produce overlapping between the two classes, which makes the classification very challenging. 
\end{itemize}

%%%%%%%%%%%%%%%%%%%%%%%%%%%%%%%%%%%%%%%%%%% Cricket data

\typicalfig{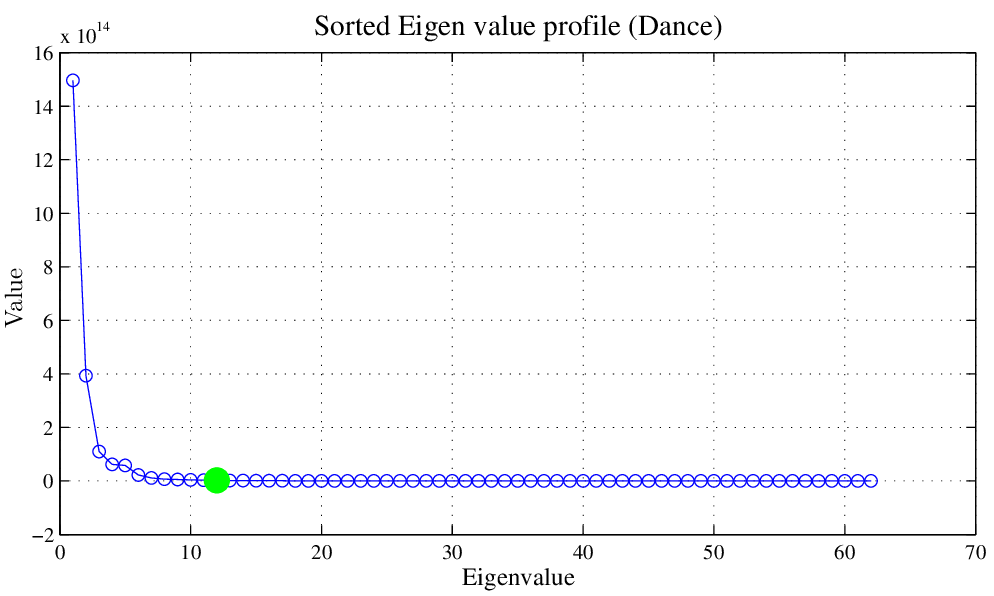}
{Eigenvalue profile sorted according to the size of the eigenvalues of matrix $\mathbf{L}$. The green circle indicates the selected dimension as the effective dimension for the regularised coefficients.}
{dance_diag_val}

\subsection {Cricket Umpire's Signals}
Cricket as a bat-and-ball team game is the world's second most popular sport\cite{10topsport}. 
An umpire is a person in the game who makes decisions and announces different
events going on on the cricket field. The umpire communicates using his arm movements. 
As an example, the event \lq No-Ball\rq\ is signaled by holding one arm out at shoulder height to indicate that the ball is delivered along with the player's fault(s); in order to announce the start of the last hour of the
play, the umpire taps his wrist and his watch \cite{shepherd2005bbc}.

For our classification task, we use the dataset which is provided by \cite{ko2005online}. This dataset contains $12$ umpire signals, 
each performed by four different persons in $3$ to $4$ repetitions. Data are captured via  accelerometers on the umpire's wrists while performing the signals\cite{chambers2004segmentation}. 
This way, data sequences are recorded in a 3D spatial format (with $X$, $Y$, and $Z$ coordination) providing $6$ features 
per time step for the classification task. The classification task is  to distinguish between $12$ classes of different cricket umpire's signals using the available $6$ motion features and a total number of $180$ data samples.

%%%%%%%%%%%%%%%%%%%%%%%%%%%%%%%%%%%%%%%%%%% Words
\begin{table}[bt]
	\renewcommand{\arraystretch}{1.3}
	\small
	\caption{Classification accuracy(\%) for the proposed DTW-LMNN approach and its comparison to the regular Euclidean-LMNN and DTW-kNN methods for the four data sets. Variances are reported in parentheses, and
	a paired t-test checks the hypothesis that DTW-KNN and DTW-LMNN do not differ.}
	\label{tab:AccuracyComparison}
	\centering
	\begin{tabular}{l||p{1.6cm}||p{1.3cm}||p{1.3cm}||p{1.3cm}}
		\hline
		& \bfseries Euclidean LMNN & \bfseries DTW-KNN & \bfseries DTW-LMNNN & \bfseries p-value \\
		\hline\hline
		Walking&92 (0.87) &95 (0.77) &100 (0) & -- 
			\\\hline\hline
			Dance&80 (1.49)&77.5 (1.51)&90 (1.03) & 0.01 
			\\\hline\hline
			Cricket Signals&95.56 (0.38)&99.44 (0.18)&100 (0)&-- 
			\\\hline\hline
			Words&97.30 (1.20) &98.61 (1.05) &99.06 (1.11) & $<0.01$ 
			\\\hline
		%\hline
	\end{tabular}
\end{table}

\begin{table}[b!]
	\renewcommand{\arraystretch}{1.3}
	\caption{Accuracy of the KNN classifier used for low-rank data/distance representations as obtained by PCA, low-rank Euclidean LMNN, and low-rank LMNN with DTW, choosing rank $3$.
	A paired t-test evaluates the hypothesis that the results of a PCA projection together with DTW and LMNN
	are the same as low-rank DTW-LMNN learning.}
	\label{tab:DRComparison}
	\centering
	\begin{tabular}{l||p{1.8cm}||p{1.8cm}||p{1.8cm}||p{1.3cm}}
		\hline
		& \bfseries low-rank Euclidean LMNN & \bfseries PCA with DTW-KNN & \bfseries low-rank DTW-LMNNN & \bfseries p-value \\
		\hline\hline
		Walking&86.6 (1.10)&96 (1.08)&98.8 (1.80)& $<0.01$
			\\\hline\hline
			Dance&75 (1.52)&76 (1.51)&95 (0.80)&0.012  
			\\\hline\hline
			Cricket Signals&96.11 (0.46)&99.44 (0.18)&100 (0)&-- 
			\\\hline\hline
			Words&98.60 (0.14)&94.24 (0.25)&99.12 (0.17)& $<0.01$
			\\\hline
		%\hline
	\end{tabular}
\end{table}

\subsection{Articulatory Words}
People can have oral communication difficulties based on different reasons; 
one example is the treatment of larynx cancer via surgery, after which it is probable that the patient  has 
significant vocal impairments. One practical solution to facilitate this issue is to benefit from silent \lq speech\rq\
recognition \cite{wang2013word,wang2014preliminary,wang2012whole}: this technology
uses facial data (such as lips and tongue movements) to recognize the person's voiceless uttered 
words or phrases \cite{wang2014preliminary}. The required motion information can be captured by attaching Electromagnetic Articulograph (EMA) sensors to the person's articulators such as lips and tongues \cite{yunusova2009accuracy}. 
The given task is to classify
the uttered word from this movement data reliably.
%An EMA is a small device that is able to track the movements of the important articulators with a satisfactory accuracy. As an example, the AG500 sensor has shown a bounded median error of 0.5 mm while testing with various  types of speech recording \cite{yunusova2009accuracy}. 

For our experiments, we use the articulatory dataset from \cite{wang2013word}, which consists of EMA data related to $25$ words uttered by different native English speakers. For the motion capture process, $12$ EMA sensors are used to capture the 3D information ($X$, $Y$, and $Z$ positions) of various facial organs as time series data. 
As it is explained in \cite{wang2013word}, the sensors are placed at different parts of forehead, lips, and tongues shaping up $36$ available features, out of which we use $9$ specific dimensions for our classification task collected from markers on the tongue tip, the upper lip, and lower lip. More precisely, we used the 3D spatial data ($X$, $Y$, and $Z$) related to the tip of the tongue (T1), upper lip (UL) and lower lip (LL) which results in $9$ features in total \cite{shokoohigeneralizing}. Hence the classification task can be characterized as categorizing $25$ classes of different words using $575$ samples of data which are represented based on $9$ motion-based features.  
%
%%%%%%%%%%%%%%%%%%%%%%%%%%%%%%%%%%%%%%%%%% classificaiton acuracy
%\clearpage
% Mat-files:  LMNN_DTW_walking_7_class_49_CV_20     ,  LMNN_Euc_walking_7_class_49_CV_K_20  
						% LMNN_DTW_dance_2class_35_CV						,  LMNN_Euc_dance_2class_35_CV
						% LMNN_DTW_Cricket_12_class_180_data_93 ,  LMNN_Euc_Cricket_12_class_180_data_CV 
						% LMNN_DTW_Words_25_class_575_mont      ,  LMNN_Euc_Words_25_class_575_mont
						
\subsection{Classification Accuracy}\label{sec:sec_acc}
In this section, we study the proposed methods as concerns their classification accuracies. We compare the classification accuracy of the DTW-LMNN algorithm, Euc-LMNN, and DTW-KNN. 
We accompany the resulting accuracies  by the variance, and
by  the p-value from a paired t-test testing the hypothesis that
plain DTW and the metric adaptation technique DTW-LMNN 
yield the same result \cite{seiler1989numerical}. 
All results from the classification task are displayed  in Table \ref{tab:AccuracyComparison}.

According to the classification results \ref{tab:AccuracyComparison}, DTW-LMNN outperforms the Euclidean version of the algorithm for all data sets.
This observation supports the expectation that DTW constitutes a suitable dissimilarity measure for  motion data
due to its ability to account for different motion durations. From another point of view, metric adjustment
enables an improvement of the classification accuracy for all cases.
Interestingly, it causes a slight superiority of the Euclidean metric as compared to DTW without metric adjustment for the dance dataset.
For all settings, metric adjustment leads to an improvement of 
the classification in connection to DTW.

\typicalfig{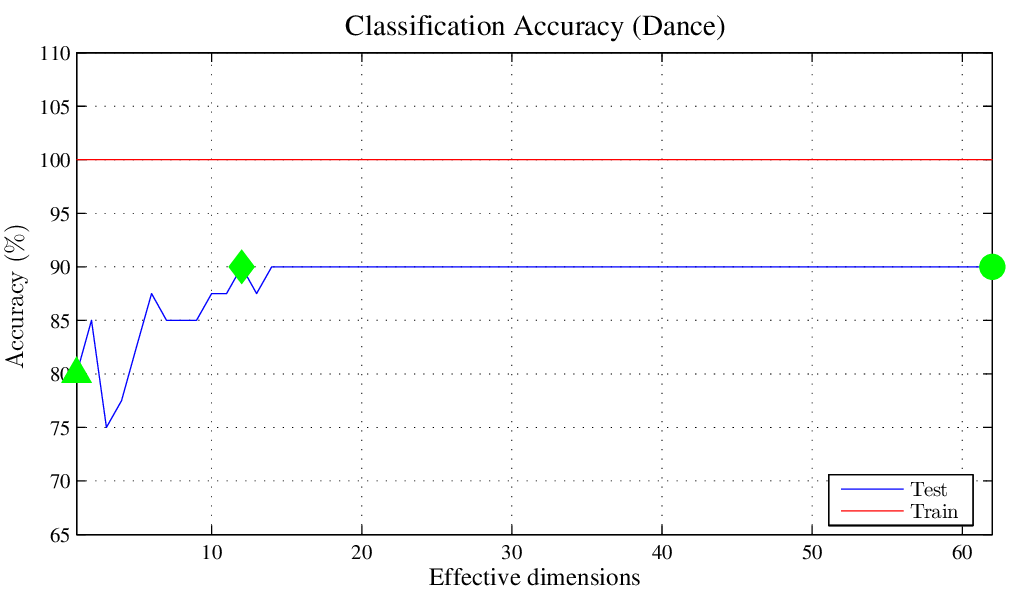}
{Classification accuracy for training and test set of the dance dataset based on regularized coefficients. The green diamond represents the highest accuracy for the test set for $12$ effective dimensions. The green circle refers to the non-regularized coefficients, and the triangle to only one effective dimension.}
{dance_eff}
%======================== Dimension Reduction
% Mat-files:  LMNN_DTW_dance_2class_35_CV_out_3_k_20	 , LMNN_Euc_dance_2class_35_CV_out_3_k_20, 
						% LMNN_DTW_walking_7_class_49_mon_out_3    , LMNN_Euc_walking_7_class_49_CV_out_3_k_20
						%	LMNN_DTW_Cricket_12_class_180_data_CV_out_3_k_10		, LMNN_Euc_dance_2class_35_CV_out_3_k_20
						% LMNN_DTW_Words_25_class_575_CV_out_3			,		  LMNN_Euc_Words_25_class_575_CV_out_3_k_10

\subsection{low-rank matrix representation}\label{sec:sec_DR}
We study the dimensionality reduction performance of DTW-LMNN  using the datasets 
introduced in section \ref{sec:dataset},
in order to obtain a compressed representation of the data for the classification tasks. As discussed in section \ref{sec:lmnn} we can use a low-rank matrix $\mathbf{M}$ or $\mathbf{L}$,
corresponding to low-rank constraints in the optimization problem (Eq.~\ref{dtwlmnn}).
Apart from a compressed representation,
this can lead to a significant increase in  the time performance  of the kNN classification in the low-dimensional projection space \cite{weinberger2008fast}. 
We use a rank $3$ matrix $\mathbf{L}$ corresponding to a projection 
into the space $\mathbb{R}^3$.  
For comparison, we also investigate the effect of a rank restriction for  the Euclidean version of LMNN,
and we investigate the result of classical PCA for dimensionality reduction of the data before classification.
The results of these low-rank classification pipelines are
reported in  Table~\ref{tab:DRComparison}.

\relfig{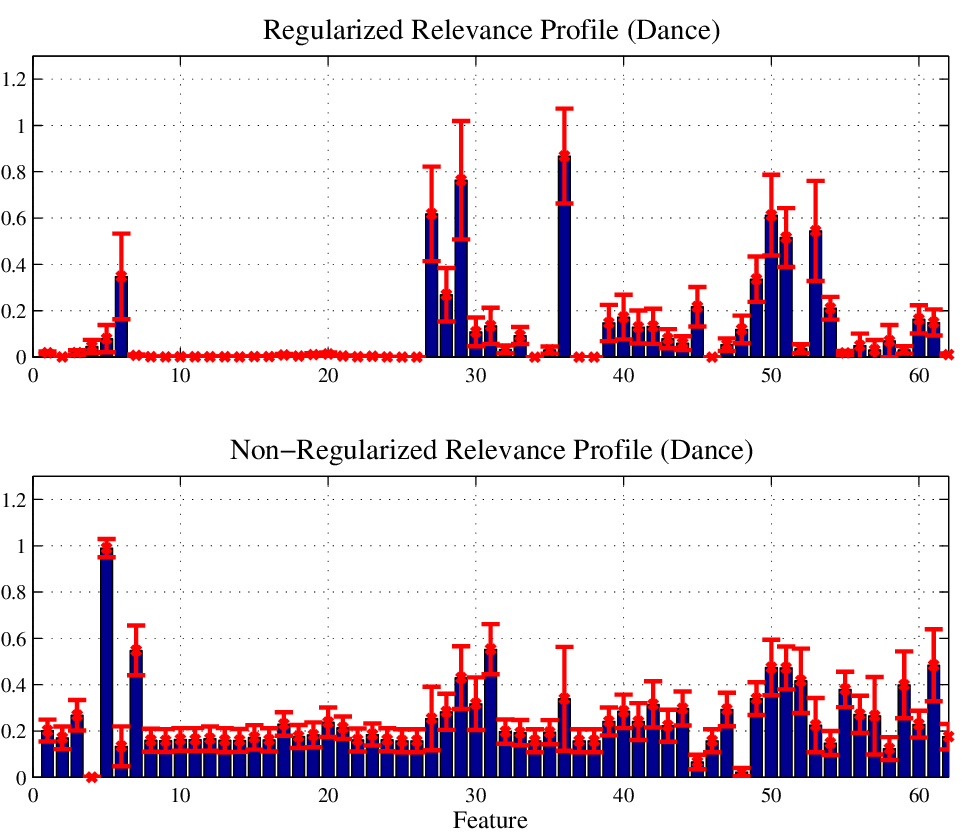}
{dance}

As reported in Table~\ref{tab:DRComparison},  low
rank DTW-LMNN preserves the good classification 
accuracy of DTW-LMNN for the reported data sets.
In contrast, PCA does not achieve the same accuracy, nor does Euclidean LMNN.
Interestingly, rank restrictions improve the classification accuracy for DTW-LMNN for the \emph{Dance} data set.
Conversely,  PCA reduces the classification accuracy for the 
\emph{Articulatory words} data set. Hence projection directions 
which are learned by LMNN optimization can enhance the discriminative aspects of DTW alignment for a low-rank matrix representation. As an interesting point in the results, the DTW-LMNN algorithm managed to classify the \textit{Cricket} dataset with $100\%$ accuracy while obtaining a compressed representation as well.

%%==============================================
%%========================= Regularized Relevance Profile
%%==============================================

\subsection{Regularized Relevance Profile:}\label{sec:sec_RegRel}
In this section, we investigate the  resulting relevance profiles for two of the introduced datasets for the
metrics obtained by DTW-LMNN. 
We restrict the analysis to DTW-LMNN due to its superior performance for all data sets.
Further, we investigate only two of the four datasets 
to focus on the notable effects of the matrix regularization.
The data sets \textit{Articulatory Words} and \textit{Cricket Signals} are of minor interest
for this section due to their comparably low-dimensionality
($9$ and $6$ sensors only, without considerable correlations). 
On the contrary, the two human body datasets (\textit{Dance} and \textit{Walking}) 
have a high number of features ($62$) with substantial correlations among  
the joints. Hence we can expect interesting effects 
when regularizing the learned matrix.

Matrix regularization has different effects:
(I) It enables a valid interpretation of the feature relevance profile
since it avoids spurious relevance peaks and random effects due to
data correlations -- we evaluate this effect by an inspection of
the  sparsity and variance of the relevance profile within cross-validation.
(II) It suggests the possible ways to reduce the data dimensionality by eliminating the
most irrelevant features according to the found relevance profile. 
We investigate this effect by evaluating the classification
performance if the feature dimensions are iteratively deleted
according to their relevance.

%\clearpage
\featselfig{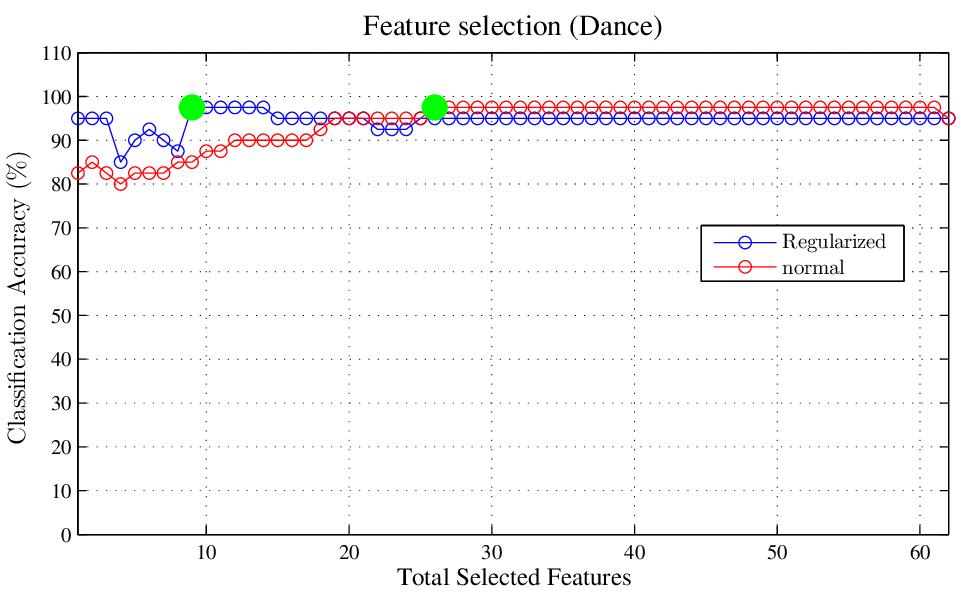}{dance}
%%==================================== Dancing
%\clearpage

\subsubsection{Dance dataset}\label{sec:rel_dance}
For the \textit{dance} dataset, we calculate the relevance value of features as $\mathrm{diag}(\mathbf{L}^t\mathbf{L})$ 
based on the transformation matrix $\mathbf{L}$ which is obtained via DTW-LMNN,
see section~\ref{sec:sec_acc}. 
For the graphical display, we normalize the profiles to the range $[0, 1]$.
Since $\mathbf{L}$ is different for different cross-validation partitions,
we report the average and variance of each diagonal value.
The resulting relevance profile without regularization 
is displayed in Fig.~\ref{fig:dance_Rel}-bottom. 
The total variance of this profile is $4.47$.

In comparison, we regularize
the matrix $\mathbf{L}$ according to Eq.~\ref{dtwtrafo}.
Thereby, the eigenvectors $\vec u^i$ of the vectors of the distance 
matrix $\mathbf{D}$ for the eigenvalue $0$ are determined based on the training set.
We report the result for the choice of $12$ effective dimensions of the eigenvectors by showing the corresponding eigenvalue profile
in Fig.~\ref{fig:dance_diag_val}.
The resulting regularized profile is shown in 
Fig.~\ref{fig:dance_eff}-top. Clearly, much fewer features are singled out as relevant.
Further, the variance of the profile is reduced to $2.86$.

\stickfig{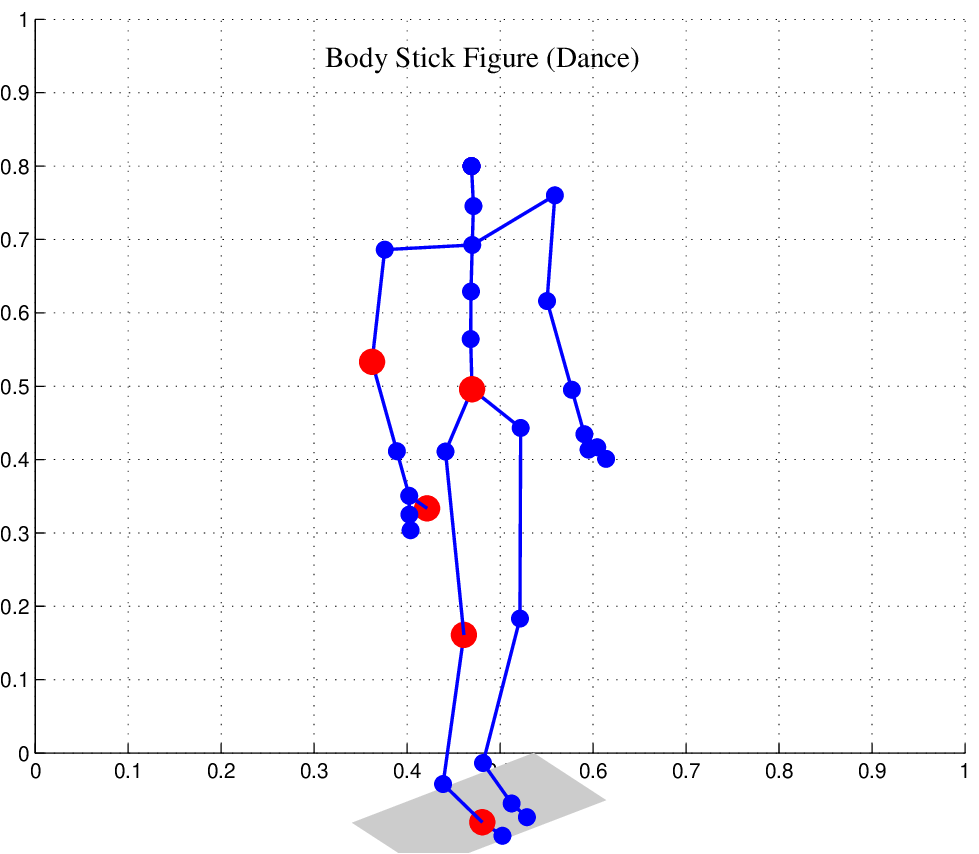}{dance} 

We test the ability of metric learning to induce a feature selection by
sorting the input dimensions (features) according to their relevance values
 in each of the two relevance profiles in Fig.~\ref{fig:dance_Rel}.
 Then we select the important features according to this order and report the resulting
 classification accuracy on the test set (Fig.~\ref{fig:dance_feat_sel}).
%Doing so, the classification accuracy as displayed in Fig.~\ref{fig:dance_feat_sel} results.
Interestingly, both relevance profiles let us to remove a large number of sensors
without a reduction of the classification accuracy. This feature reduction can be
iterated until only $9$ features are left for the regularized profile, and $26$ features for the
non-regularized one. Hence regularization greatly enhances the feature selection ability
of the technique. The resulting $9$ features are  
displayed based on the skeleton information  in Fig.~\ref{fig:dance_stick}.
The semantic meaning of these features  is reported  in Table.~\ref{tab:dance_regul}.

\begin{table}[b!]
	\renewcommand{\arraystretch}{1.3}
	\caption{Total variance in the regularized and non-regularized relevance profiles along with the feature selection results for the \textit{Dance} dataset. }
	\label{tab:dance_regul}
	\centering
	\begin{tabular}{l}
		\hline
	{\bfseries  Total variance of profile (before regularization)}: 4.47 \\
		\hline\hline
		{\bfseries  Total variance of regularized profile}:  2.86 \\
		\hline\hline
		{\bfseries Selected Joints and related feature number:  }  rhumerus(27,28,29),\\
		 rthumb(36), 
	 rfemur(49,50,51), rfoot(53) and root(6) \\
		\hline
		%\hline
	\end{tabular}
\end{table}

According to Fig.~\ref{fig:dance_Rel}, the regularization matrix has positive effects on the relevance profile. 
While it retains the classification performance at its highest rate (see Fig.~\ref{fig:dance_eff}), it reduces redundancy in the profile and produces a sparse representation for the relevance values of inputs (features). Besides, based on the variance measure for the relevance profile over different cross-validation partitions, the regularized profile has less variance and thus is more reliable than the normal one.

As a semantic interpretation, it can be concluded from Fig.~\ref{fig:dance_stick} that \textit{hands} and \textit{feet} are both important discriminative features for this dancing task. From another point of view, this is a difficult task because each class has different subcategories within itself which account
for overlaps with other class; hence the combination of both (hand and foot) is required to distinguish between the two dance categories. Furthermore, as another interesting semantic interpretation, only the data related to one side of the body (right side) is necessary to achieve the highest classification performance. 
This interpretation coincides with the fact that dancing is typically a symmetrical whole body movement in which symmetry can be found between the left and right sides of the body.

%%==================================== Walking
%\clearpage
\subsubsection{Walking dataset}
We repeat this experimental setting for the \textit{walking dataset},
which leads to $14$ effective dimensions selected from the regularization matrix. 
The obtained regularized profile can be seen in Fig.~\ref{fig:Walking_Rel}.
The total variances of the relevance profiles before and after the regularization are $10.7$ and $2.51$, respectively. Again, similar to the dance dataset, the most relevant features, and the resulting classification accuracies are displayed in
Fig.~\ref{fig:walking_feat_sel}.
The essential joints are listed in Tab.~\ref{tab:walking_regul}.
%\\
%\textbf{Selected Joints/Features:} root(5), lhumerus(40,41), rhand(33), rthumb(36), lowerneck(18) and lthumb (48).
%\\
%\textbf{Sorted Selected features ($Descending$):} 5 ,   41 ,   33,    36   , 40   , 18   , 48

\relfig{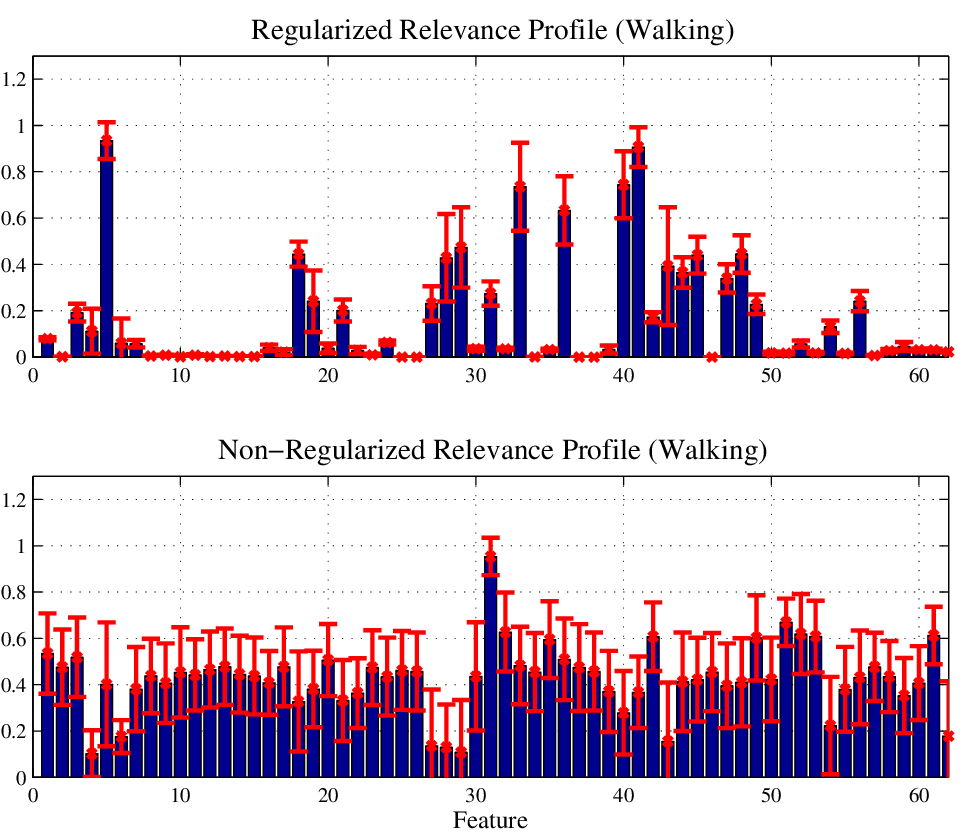}
{Walking}

%\relfig{7-Walking_CV_Rel_bar.eps}
%{Walking}

%\typicalfig{7-Walking_CV_Rel_bar.eps}
%{Mean and Deviation of "Relevance value" for \textbf{Walking dataset} for all features according to diagonal value of $L^TL$}
%{Walking_Rel}

%\clearpage

\begin{table}[b!]
	\renewcommand{\arraystretch}{1.3}
	\caption{Total variance of the regularized and non-regularized relevance profiles 
	and selected features for the \textit{Walking} dataset. }
	\label{tab:walking_regul}
	\centering
	\begin{tabular}{l}
		\hline
	{\bfseries  Total variance of plain profile}: 10.70\\
		\hline\hline
		{\bfseries  Total variance of regularized profile:}  2.51\\
		\hline\hline
		{\bfseries Selected Joints and feature number}: root(5),
		 lhumerus(40,41), \\ rhand(33), 
 rthumb(36), lowerneck(18) and lthumb (48) \\
		\hline
		%\hline
	\end{tabular}
\end{table}

Similar to the results  for the dance dataset, regularization of the learned metric results in a sparse representation of the relevance profile and a reduced variance.
Furthermore, according to Fig.~\ref{fig:walking_feat_sel}, a classification accuracy of $100\%$ 
can be achieved while choosing fewer features ($7$ features for
the regularized profile instead of $25$ for the standard one). 

Based on the observations from Fig.~\ref{fig:walking_stick}, for this dataset (and this classification task), hands are more  important than feet. In addition, as the classes are very similar
(all of them are connected to \textit{walking}), the classification algorithm also needs some information about both sides of 
the body in order to carry out the classification task with a perfect result.
We tested this hypothesis by using   \textit{Lhand} instead of \textit{Rhand} or deleting   \textit{Rthumb} (since we already have \textit{Lthumb}), but in both cases, the performance decreased (around $3$ to $4\%$) showing that those selected features are all necessary even though they are symmetrical in the skeleton structure.

%======= results

According to the results given in section \ref{sec:sec_RegRel}, after applying the regularization matrix,
 both datasets showed improved results. 
The regularization enhances the reliability of the relevance profiles, corresponding to
a smaller variance. 
In addition, regularization enables a more efficient feature selection strategy.

\featselfig{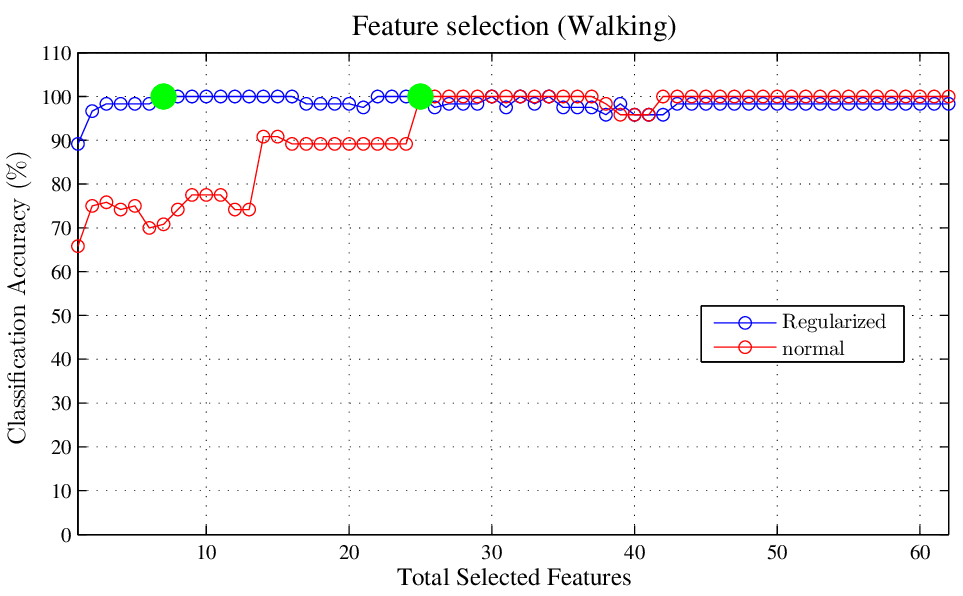}{walking}

\stickfig{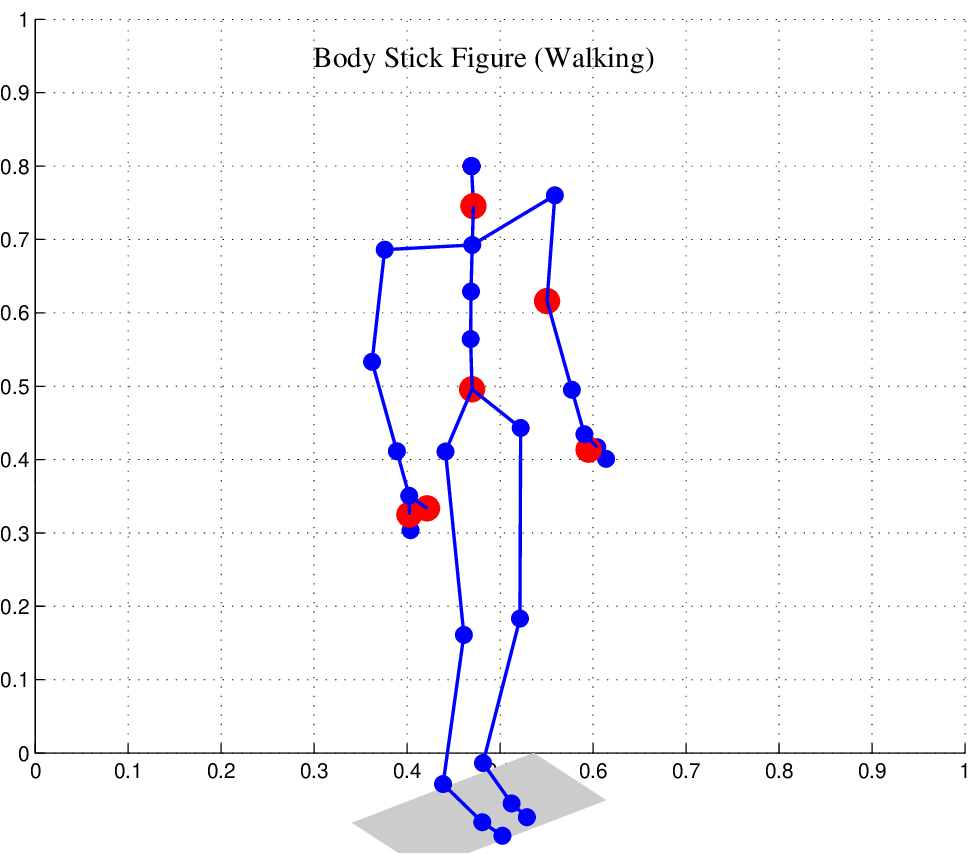}{walking}

%TODOs
%%%%%%%%%%%%%%%%%%%%%%%%%%%% Conclusion
%\clearpage
\section{Conclusions and Future Work}
In this paper we introduced a distance based extension (DTW-LMNN) to the popular metric learning method LMNN in a very generic way, opening up the possibility to also transfer auxiliary concepts such as metric regularization. 
This framework enables us to benefit from the DTW dissimilarity measure which
is particularly suited to the analysis of Mocap data. While dealing with multi-dimensional motion data such as human movements, the component-wise dissimilarity values achieved by the DTW measure can be treated as different features to obtain a semi-vectorial representation for the dissimilarities among the data. 
Therefore, the representation can be combined with the LMNN framework to efficiently adapt the
feature ranking and correlation according to the classification tasks at hand. The mentioned dissimilarity representation for the motion data also conserves the convexity of the optimization problem Eq.\ref{lmnn}. 

The strong results achieved with DTW-LMNN show that augmenting the LMNN approach with the DTW dissimilarity measure can significantly improve the metric learning performance. For both, classification and dimensionality reduction tasks, the proposed approach outperforms the Euclidean version of LMNN as well as the DTW based K-nearest neighbor method. Therefore, it can be concluded that DTW-LMNN can benefit from the strength of the DTW dissimilarity metric and LMNN metric learning together. According to our encouraging results across diverse motion-based benchmarks, the DTW-LMNN framework offers a suitable discriminative method to achieve high-performance classifications and compressed representations dealing with motion-based datasets. 

As another contribution of this paper, we devised a way to transfer the concept of metric regularization to alignment based representations; this concept 
has recently been proposed for the vectorial case\cite{l2reg}. According to the results in section \ref{sec:sec_RegRel}, this regularization step is a crucial prerequisite for a valid interpretation of the relevance profile. To that end, we managed to use the dissimilarity-based information in order to remove the highly correlated dimensions related to the null space contributions. According to the results, the dissimilarity-based regularization brings significant effects to the relevance profile; further, it increases the semantic interpretability of the resulting discriminative models. It is important to mention that the proposed regularization step can be applied to any other dissimilarity-based metric framework as well.

Relying on these promising results achieved by the proposed DTW-LMNN framework, there is considerable potential for future research in dissimilarity-based metric learning: the principle can be transferred to
other metric learning methods which are not linked to KNN.
 In addition to that, a promising research line could be to investigate more advanced regularization techniques to achieve a further enriched relevance profile, such as proposed in \cite{ l1reg}.

%\clearpage
%Abstract—We investigate metric learning in the context of
%dynamic time warping (DTW), the by far most popular dissim-
%ilarity measure used for the comparison and analysis of motion
%capture data. While metric learning enables a problem-adapted
%representation of data, the majority of methods has been proposed
%for vectorial data only. In this contribution, we extend the
%popular principle offered by the large margin nearest neighbors
%learner (LMNN) to DTW by treating the resulting component-
%wise dissimilarity values as features. We demonstrate, that this
%principle greatly enhances the classification accuracy in several
%benchmarks. Further, we show that recent auxiliary concepts such
%as metric regularization can be transferred from the vectorial
%case to component-wise DTW in a similar way. We illustrate,
%that metric regularization constitutes a crucial prerequisite for
%the interpretation of the resulting relevance profiles.

%\clearpage
%\section{Acknowledgment}
%This research was supported by the Cluster of Excellence Cognitive 
%Interaction Technology 'CITEC' (EXC 277) at Bielefeld University, which
%is funded by the German Research Foundation (DFG).

	%\section{Acknowledgment}
	%This research was supported by the Cluster of Excellence Cognitive 
	%Interaction Technology 'CITEC' (EXC 277) at Bielefeld University, which
	%is funded by the German Research Foundation (DFG).

	%\begin{footnotesize}
	\newpage
	\bibliographystyle{unsrt}
	%\def\bibfont{\tiny}
	%\tiny
	%\scriptsize
	%\bibliography{C:/Thesis/Publications/Ref4Papers_CS}
	%\bibliography{IEEEabrv,Ref4Papers_CS,bibs}
	\bibliography{IEEEabrv,/vol/semanticma/Thesis/Publications/Ref4Papers_CS,bibs,Ref4Papers_CS}
	%\bibliography{Ref4Papers_esann}
	
	%\end{footnotesize}

\end{document}